# Object category understanding via eye fixations on freehand sketches

Ravi Kiran Sarvadevabhatla, Sudharshan Suresh, and R. Venkatesh Babu, *Senior Member, IEEE*

*Abstract*—The study of eye gaze fixations on photographic images is an active research area. In contrast, the image sub-category of freehand sketches has not received as much attention for such studies. In this paper, we analyze the results of a free-viewing gaze fixation study conducted on 3904 freehand sketches distributed across 160 object categories. Our analysis shows that fixation sequences exhibit marked consistency within a sketch, across sketches of a category and even across suitably grouped sets of categories. This multi-level consistency is remarkable given the variability in depiction and extreme image content sparsity that characterizes hand-drawn object sketches. In our paper, we show that the multi-level consistency in the fixation data can be exploited to (a) predict a test sketch's category given only its fixation sequence and (b) build a computational model which predicts part-labels underlying fixations on objects. We hope that our findings motivate the community to deem sketch-like representations worthy of gaze-based studies vis-a-vis photographic images.

## I. INTRODUCTION

WHEN shown photographic images under a free-viewing (i.e task-free) paradigm, human eyes preferentially fixate on image locations which are visually salient. Multiple studies [1]–[5] have demonstrated that this fixation mechanism is bottom-up, predominantly driven by image content and richness of detail (color, texture etc.).

This explanation, while satisfactory for photographic images, seems inadequate for certain categories of images such as line drawings. In particular, one class of line drawings – hand-drawn sketches – are sparse and largely devoid of detailed content. In addition, they are typically binary images containing virtually no color-based information (see Fig. 1). Even so, multiple studies have demonstrated a "fixations-into-nothing" phenomenon [6]–[9], wherein the eye fixations on the same stimulus by multiple subjects fall on empty regions, yet exhibit enough regularity to make gaze-based inferences. One possible explanation is that the first eye fixation conveys all there is to know ('Gestalt') about the underlying scene semantics [10] and the regularity in rest of the fixations is a statistical anomaly. However, a more intriguing explanation is that these empty region fixations aim to implicitly verify the overall consistency of the scene content depicted in the sketch [11], [12]. Which of these explanations is correct?

On a separate note, gaze-tracking studies have demonstrated that photos of objects trigger signature gaze patterns and go

Ravi Kiran Sarvadevabhatla (ravika@gmail.com) and R. Venkatesh Babu (venky@cds.iisc.ac.in) are with Department of Computational and Data Sciences, Indian Institute of Science, Bangalore 560012, India. Sudharshan Suresh was a summer intern at Video Analytics Lab.

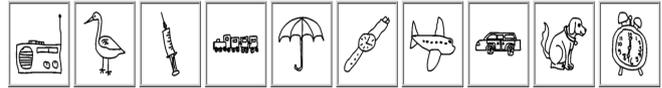

Fig. 1: A sampling of the freehand sketches from the dataset of Eitz et al. [14] used in our eye-fixation user study.

as far as to say that the underlying object category can be predicted from gaze patterns alone [2], [8]. Do hand-drawn sketches of objects elicit gaze patterns which are predictive of the object category as well? Suppose the category of the sketch is made known shortly prior to actual viewing of the sketch ('priming'). Do the gaze patterns exhibit correlation with any object-level attributes such as semantic-parts? [13].

To examine these questions and related issues, we conducted an eye fixation study on a large database containing hand-drawn sketches of objects across 160 categories [14]. In this paper, we present some of the interesting findings from our study and discuss the larger implications of our study for visual object category understanding. We summarize our contributions and major findings below :

- We provide our database – SketchFix-160 – of eye fixations for 3904 hand-drawn sketches across 160 visual object categories (Section III).
- We use SketchFix-160 to create ground-truth eye fixation maps for object sketches. The large quantity of our maps (3904 compared to the previous largest set of 200) can be used in benchmarking generalized[1] saliency prediction models [15]–[17].
- The eye fixation sequences exhibit significant category-level structure (Section VI). In fact, we find this structure to be consistent enough to predict the sketch's category from its fixation sequence alone (Section VII), without recourse to sketch stroke data.
- We map eye fixation sequences to semantic object-part label sequences by utilizing part contour annotations of sketches [18]. Our analysis of these label sequences shows that the sequencing of fixations corresponds to an implicit visitation of the sketch object's parts in a consistent, pre-defined order of importance. This result is quite remarkable, considering that the parts themselves are not explicitly delineated in the sketch stroke data (Section VIII).
- Leveraging the part-level consistency of eye fixation sequences, we build object-specific computational models which predict the semantic object-part underlying an eye fixation (Section IX)

---

[1]These approaches aim to predict saliency for multiple types of images, including non-photographic varieties such as sketches studied in this paper.

From an application point of view, sketch-like object depictions tend to be used as stylized representations of the underlying category in 3D model renderings [19], art displays, graphic icons and product logos. Therefore, eye-fixation data can provide insights into factors which affect the way such representations are perceived. As an illustrative example, we provide a category-level analysis based on fixation density (see Section VI and Figure 5). In other work, Reid et al. [20] use eye-gaze data to interpret customer selections among sketch-like consumer product renderings. Moreover, sketch-like representations often represent a simplification of visual content. This helps constrain the number of design variables to be studied for statistical significance, potentially reducing the complexity of the study design [20].

Our sketch object eye-fixation data lays the ground for future studies which can uncover connections between eye fixation patterns on objects in photographic images and their sketched versions. Such connections can help understand depiction invariant attributes and aspects of visual object representations. Combined with data from similar studies on other modalities [13], our sketch gaze data can contribute towards new insights into cross-modal (photo, brush art, line drawing) object representations [21]. To the best of our knowledge, current theories and computational models of saliency, which implicitly assume photographs or real-world scenes as input, seem insufficient to explain our findings. Therefore, further investigation into our findings could lead to more general computational models of human visual saliency which can explain fixations regardless of depiction (photos or sketches).

Please visit http://val.cds.iisc.ac.in/sketchfix to access our SketchFix-160 dataset and additional material.

## II. RELATED WORK

The study of eye-fixations on images is a very active research area in the inter-related fields of neuroscience and computer vision [1]–[5], [22]. However, the broad image category containing sketch-like depictions has not received as much attention presumably due to lack of usable image content with which to correlate the fixation data. Within the literature available, there are four broad categories of studies involving eye-gaze tracking and sketch-like depictions. The first category of studies use gaze tracking to understand how people copy and draw line-drawings and simple shapes [23], [24]. The second use gaze tracking to study differences between perception of photographic image content and corresponding line-drawing representations [25], [26]. In the third category, gaze tracking is used to characterize semantic plausibility of objects in line-drawings of scenes [27]. The fourth category of studies use gaze tracking to explore sketch-like depictions of objects [7]–[9].

Saliency-related studies typically involve photos of scenes. Although objects make the scenes meaningful in many instances [28], [29], their role has been studied in a limited context. To address this, the role of category-level and semantic-level information has been examined on a large database of objects by Xu et al. [13]. Additionally, in their survey paper, Frintop et al. [30] summarize prevailing theories of the relationship between attention and object recognition.

Recently, Ali and Itti [16] constructed a large-scale dataset CAT2000 consisting of eye-fixation data for 20 categories of scenes. One of the 20 categories, 'Sketches', contains fixations for sketch images originally belonging to the same sketch dataset we have used [14]. However, the number of sketches in CAT2000 is quite small (200) compared to what we have studied (3904).

Our work belongs to the fourth category mentioned above (viz. using gaze tracking to explore sketch-like depictions of objects). However, we analyze a much larger pool of object sketches. To the best of our knowledge, we are also the first to analyze the sketches in terms of categories and that too, across a relatively large number of categories.

## III. DATA GATHERING PROTOCOL

For our analysis, we used 3904 sketches spread across 160 object categories studied in the work of Sarvadevabhatla and Babu [31]. In their work, the authors use sketches correctly classified by a deep-feature classifier to construct sparsified yet recognizable versions of the sketches which they term *category-epitomes*. In our study, we utilize the original, full sketches from the dataset of Eitz et al. [14] used to construct epitomic versions. Our choice of sketch data was motivated by two factors: (a) enable collection and analysis of eye-fixation data for freehand sketches across a large number of object categories while keeping the burden of data collection manageable (b) enable eye-fixation based analysis of *epitome*-like sparsified representations in future.

**Equipment and Setup:** All subjects were comfortably seated at a viewing distance of 60 cm from a 19-inch LCD monitor. The stimulus images (sketches) were effectively displayed at a resolution of $1024 \times 1024$ and centered within the display. The stimuli were all the same size. Eye tracking was performed non-invasively via real-time video feed provided by an iView X$^{\text{TM}}$ Hi-Speed system. The equipment was operated in monocular mode at a sampling rate of 500Hz. The subject pool consisted of 36 people of both genders aged between 18 and 45 years. All subjects displayed normal or corrected-to-normal vision.

For our experiments, we used 3940 sketches spread across 160 object categories. There were 24 sketches for each category and the mean length of fixation sequences was 9. The eye fixation data was collected in two regimes which we refer to as 'Unprimed' and 'Primed'.

**Unprimed Regime:** The total set of 3904 images (spread over 160 object categories) was first divided randomly into 16 groups of 244 images each. Each subject viewed sketches from a single group. These groups were shown to 20 subjects. In total, the $20 \times 244 = 4880$ image viewings in the unprimed regime resulted in a total of 42943 eye-fixations. The sketches in each group were shown to the corresponding subject in a randomized fashion. In this regime, each sketch in a group was displayed for 3 seconds[2] followed by a gray screen display

---

[2]In our pilot studies, we found that a duration above 3 seconds between stimuli resulted in pilot subjects reporting eye strain and inability to fully complete viewing our pre-requisite number of sketches per participant. With a view to minimize participant burden, we chose a presentation time of 3 seconds.

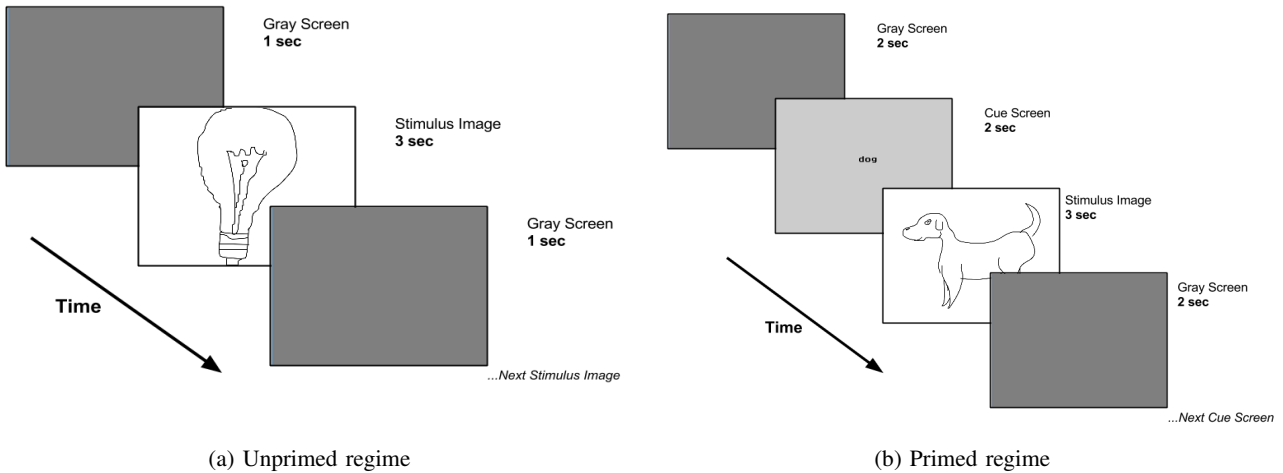

(a) Unprimed regime  (b) Primed regime

Fig. 2: An illustration of the experimental procedure for the primed and unprimed regimes in our study.

lasting 1 second, before the next sketch was shown (see Fig. 2a).

**Primed Regime:** As with unprimed regime, the images were again divided into groups. However, the images were chosen only from 13 selected categories for which semantic object-part annotations exist [18]. Thus, the effective total number of images was 283. The images were divided into 4 groups (3 groups with 71 images each and 1 with 70 images). These groups were shown to 16 subjects. Effectively, each unique image was seen by 4 subjects. In this regime, the distinguishing feature was a cue screen (which preceded the sketch) displaying the sketch's category as a text string representing the category label as provided by the creators of the freehand sketch dataset [14]. All the categories were coarse-grained. For sketch of a dog, the category label provided was dog – thus, it was neither too coarse ('animal') nor too fine-grained ('Labrador'). Timing considerations were determined by feedback from pilot studies[3]. The cue, stimulus image and a gray screen were displayed (in that order) for a period of 2, 3 and 2 seconds respectively (see Fig. 2b).

The participants were instructed to observe and freely view the images. The participants were informed about the duration for which image/gray screen would be visible. Depending on the regime, the participants were informed whether or not the object category cue would be provided. To minimize visual strain, the subjects were allowed a total of 5 breaks in the unprimed regime and 2 in the primed version. The eye tracking device was recalibrated when resuming after each break to ensure good eye tracking accuracy. All of these aspects were informed to the participants before commencing the data collection procedure. The subjects were provided details of the study and informed consent was obtained from them prior to their session.

The regimes mentioned above are motivated by the possibility that the 'Unprimed Regime' necessitates visual exploration for sketch understanding whereas the 'Primed Regime' only involves visual verification of the category label provided during the study. This distinction is crucial. In fact, we shall soon see that primed gaze sequences are more structured (Section V) and help validate the consistency of implicit object part visitation (Section IX) for sketches.

[3] For word presentation, feedback from pilot studies indicated that 1 second to be too small for the viewer to read the category label satisfactorily and 3 seconds to be far too long.

## IV. CONSTRUCTING THE FIXATION MAP FOR A SKETCH

Fixation maps summarize the fixation information averaged over multiple subjects viewing the same image (sketch) and are used as ground-truth for evaluating saliency prediction methods [1], [3], [15], [16]. We construct fixation maps for sketches based on the approach described by O'Connell et al. [2] which we summarize next.

For a given sketch $S$, let $\{(x_f, y_f)\}, f = 1, 2, \ldots N$ denote the set of fixation locations (i.e. the combined set of all fixations aggregated over all the subjects who viewed the sketch). Let $\{t_f\}$ denote the set of corresponding fixation durations. The fixation map is initialized as the sum of impulse functions centered at each fixation location and weighted by the fixation duration. This map is then convolved by a Gaussian kernel whose standard deviation is set to $1°$ of visual angle ($\sim 36$ pixels in our case) to approximate the uncertainty in eye-tracker's fixation measurements.

$$F'(x,y) = \frac{1}{\sum_{f=1}^{N} t_f} \sum_{f=1}^{N} t_f \, exp\Big(\frac{-(x_f - x)^2 - (y_f - y)^2}{\sigma^2}\Big) \quad (1)$$

Typical approaches normalize $F'$ with respect to the maximum value or to lie between 0 and 1 [1], [16]. In contrast, O'Connell et al. [2] suggest standardizing $F'$ to obtain a zero mean, unit standard deviation map $F$. Letting $\bar{F}'$ and $\sigma_{F'}$ denote the mean and standard deviation over the spatial locations of the stimulus region respectively, we have:

$$F(x,y) = \frac{F'(x,y) - \bar{F}'}{\sigma_{F'}} \quad (2)$$

Thus, we obtain the standardized fixation map $F$ corresponding to the sketch $S$. Next, we shall see how a fixation-map based similarity measure can be used to analyze the level of agreement between fixation sequences of subjects viewing a given sketch.

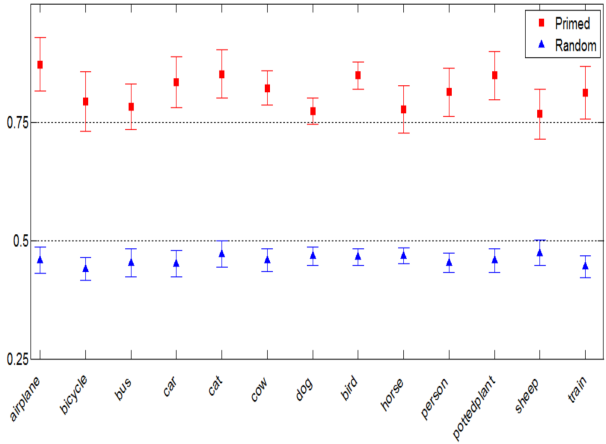

Fig. 3: Category-wise IOC (y-axis) for sketches in primed regime (in red (top)). Corresponding random IOCs are in blue (bottom). The sAUC-based IOCs are high and well-separated from random IOC in all the categories. Best viewed in color.

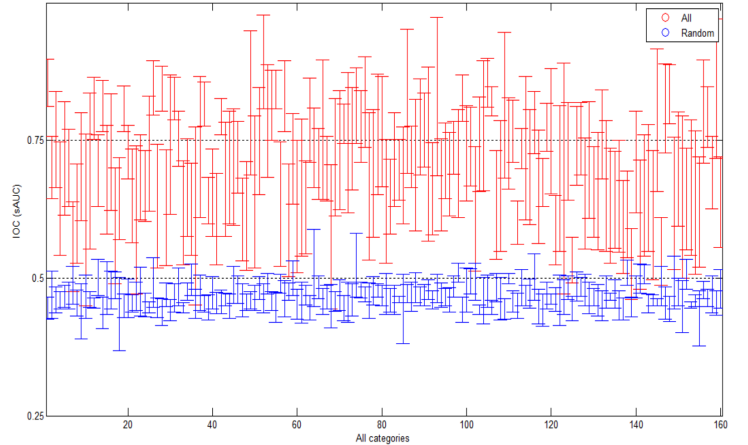

Fig. 4: Category-wise IOC (y-axis) for all sketches (in red (top)). Corresponding random IOCs are in blue (bottom). The category-wise IOC is still high, although separation from random IOC is not as distinct compared to primed categories' IOC (Figure 3). Best viewed in color.

## V. INTER-OBSERVER CONGRUENCY (IOC)

Suppose a given sketch $S$ is viewed by $N$ subjects $s_1, s_2, \ldots s_N$ resulting in $N$ fixation sequences $F_1, F_2, \ldots F_N$. To measure the degree of agreement between the subjects, we compute the Inter-observer congruency (IOC) by computing the similarity $\mathcal{D}(F_i, F_{[1:N]\setminus i})$ between fixations of subject $s_i$ and the set of fixations corresponding to the rest of the subjects ($F_{[1:N]\setminus i}$). The IOC value for sketch $S$ is computed as the average of similarity values over all the subjects, i.e. $IOC_S = \frac{1}{N}\sum_{j=1}^{N} \mathcal{D}(F_j, F_{[1:N]\setminus j})$. For our experiments, we use shuffled AUC (sAUC) as the similarity measure $\mathcal{D}$. To compute sAUC, the set of fixations from rest of the subjects (i.e. $F_{[1:N]\setminus i}$) are utilized to construct a 'saliency map' using the procedure described in Section IV. This saliency map is thresholded and used as a binary classifier to predict saliency of fixations $F_i$ from the reference subject. The resulting true and false positive rates at various threshold settings are used to compute the final similarity. Please refer to Zhang et al. [32] for details of sAUC computation.

For each category, we first compute IOC score for each sketch. Next, we compute the median across IOC scores of all the sketches within the category. We compute two groups of IOC scores. The first group comprises of IOC scores obtained using only the fixation sequences collected in the primed regime (Section III). The second group comprises of IOC scores across all the sequences regardless of regime (primed and unprimed). To verify the statistical significance of IOC scores (i.e. whether they exhibit above-chance similarity), we also compute IOC between each subject's fixation sequence and randomly generated sequences. We find the median of the resulting IOCs across sketches and refer to the same as random median IOC of the category.

A high value for IOC indicates consistency among the fixation sequences and is commonly observed for natural images, particularly with a central object [1]. The results for the first group of IOC scores (primed regime) can be seen in Figure 3. Across the categories, the median IOCs (in red) are seen to be high and well-separated from the corresponding random median IOCs (in blue). This trend is repeated for the second group of IOC scores (Figure 4) across all the 160 categories, albeit with slightly decreased median IOCs and relatively smaller separation from the random median IOCs compared to the primed regime. The high median IOCs indicates the overall reliability of our SketchFix-160 dataset.

In the next section, we move beyond per-sketch fixations and analyze fixations at a category level. To enable this analysis, the data from fixation sequences of all the sketches from a given category is combined into a category-specific spatial map. We start by describing the construction of this category-level map.

## VI. FIXATION MAPS : A CATEGORY-LEVEL PERSPECTIVE

To obtain a category-level perspective of fixations, the category-specific fixation maps $U_c$ are computed by averaging the standardized fixation maps of sketches within a category (Equation 2). Let $n_c$ denote the number of fixation maps in category $c$. We have:

$$U_c = \frac{1}{n_c}\sum_{i=1}^{n_c} F_i \quad (3)$$

Additionally, to compensate for biases common to all the categories (e.g. center-bias), marginalized versions of the category maps are computed by subtracting each category's map from the average over all the category-level maps, i.e.

$$M_c = U_c - \frac{1}{C}\sum_{j=1}^{C} U_j \quad (4)$$

In the above procedure, standardization helps highlight locations which are statistically fixated more than average

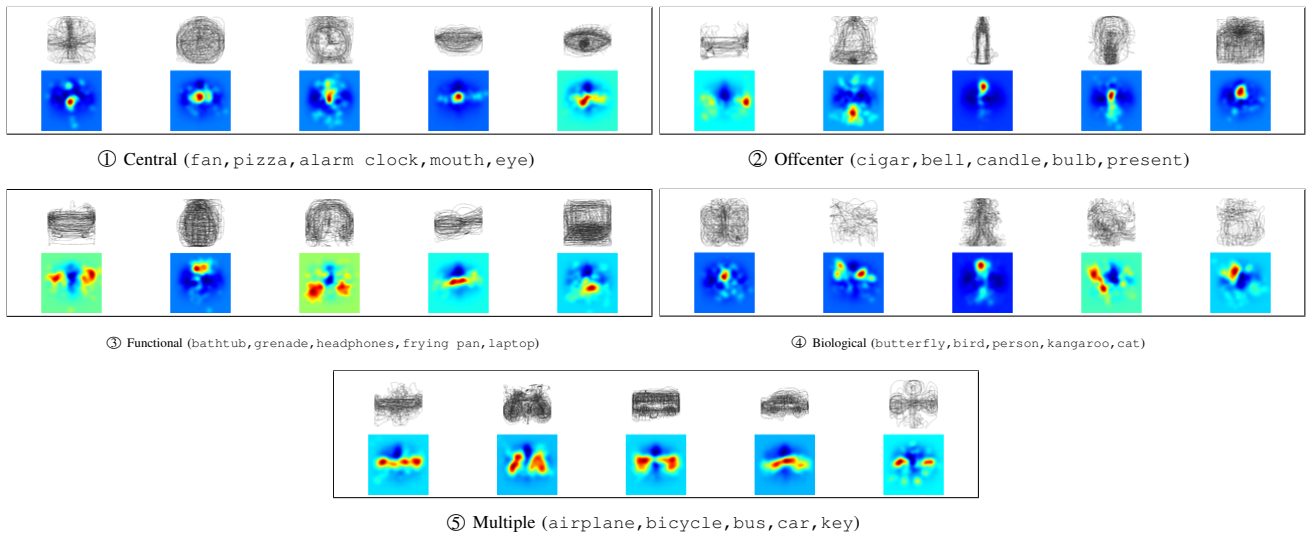

Fig. 5: Category groupings based on spatial distribution and multiplicity of hotspots in category-level fixation maps. In each group, there are two images for each category. The top image corresponds to a 'category sketch' created by adding all sketches of the category. The bottom image is the category's fixation map. Best viewed in color.

("hotspots") or less than average across instances[4]. Next, we shall visualize the category-level maps created in this manner.

**Trends in category-level maps:** An analysis of the multiplicity and spatial distribution of highly-fixated regions ("hotspots") in category-level maps reveals certain interesting groupings among the categories. Some of the categories have a sharp, centrally dominant hotspot. Visually, the sketches of these categories tend to depicted symmetrically and compactly (See bottom row of panel ① in Figure 5).

In some of the categories, the location of the hotspot is off-center. Among such categories, three groups were observed. In the first group (panel ② in Figure 5), the hotspot is the location associated with activity or potential for activity (e.g. the glowing end of cigarette, candle or the gong of a bell). Categories in the second off-central hotspot group tend to be, borrowing the terminology of Yun et al. [33], "functional" categories (i.e. crafted for a specific purpose). For these categories (panel ③ in Figure 5), hotspots tend to be on the portion of the object which is typically associated the most with the underlying category's functionality (e.g. the taps of the bathtub, the detonation pin in the grenade, the earpads of headphone etc.). The third group corresponds to "biological" categories (panel ④ in Figure 5) wherein the hotspot is predominantly located on the head or face of the biological entity. This phenomenon has been repeatedly observed for regular image datasets as well [1], [16].

Some of the categories tend to have multiple hotspots (panel ⑤ in Figure 5). Interestingly, most of the vehicular categories (airplane, bicycle, bus, car) seem to exhibit this multi-hotspot pattern.

The groupings mentioned above are not disjoint – for instance, a category in 'Offcenter' group could also be a member of the 'Functional' group (e.g. cigarette). Nevertheless, the hotspot-based structural grouping of categories demonstrates that eye-fixations on sketches are not merely a random after-effect of the underlying objects' gestalt. In fact, the statistical structure of the fixation sequences is sometimes sufficient for predicting the underlying category, as we shall see next.

## VII. OBJECT CATEGORIZATION: HOW WELL CAN WE PREDICT CATEGORY FROM FIXATIONS ALONE?

The high level of IOC observed in Section V suggests a reliable consistency among fixation sequences and raises an intriguing question : Given a fixation sequence, is it possible to predict the category of the sketch it corresponds to, just from fixation locations alone ? In this regard, a good prediction performance would enable finer, object-part based analyses which rely on a correct prediction of the underlying category (See Section IX).

To answer the question posed above, we follow the procedure of O'Connell et al. [2]. We hold out the fixation data of one subject for testing and utilize the fixation data from the rest of the subjects (Leave-One-Subject-Out method) to build category-wise fixation maps as described in Section VI. Suppose the subject whose data has been earmarked for testing has viewed $M$ sketches. Let $\{(x_f^{(j)}, y_f^{(j)})\}, f = 1, 2, \ldots N_j$ denote the set of fixation locations for the $j$-th sketch ($j = 1, 2, \ldots M$). Let $\{t_f^{(j)}\}$ denote the set of corresponding fixation durations. For a given category $c$, the prediction score is computed from the corresponding marginalized fixation map $M_c$ (Equation 4) as:

$$G_c(j) = \frac{\sum_{f=1}^{N_j} t_f^{(j)} M_c(x_f^{(j)}, y_f^{(j)})}{\sum_{f=1}^{N_j} t_f^{(j)}} \quad (5)$$

The category prediction for the $j$-th sketch is then computed as $c^*(j) = \underset{c}{\operatorname{argmax}} \ G_c(j)$

---

[4]Instances can refer to fixation locations within a single sketch or within a category

The prediction procedure is repeated for all the $M$ sketches viewed by the test subject. The resulting predictions are aggregated category-wise. The overall prediction rates are obtained by averaging the category-wise predictions across the test subjects. The average category-wise prediction rates of the Leave-One-Subject-Out procedure described above can be viewed in Figure 6 for the primed regime. The performance for most categories well exceeds chance (reciprocal of number of primed categories). To determine the effect of durations, we also obtained prediction rates for the condition where duration is ignored (by setting all durations to 1). We found that inclusion of duration information results in a slight performance improvement overall. The inclusion of duration in saliency map construction provided approximately a 5% average improvement in category prediction accuracy for primed regime (Cohens d = 0.54, medium effect). Specifically, the median prediction rate across the categories was 32% (ignoring duration) and 37.2% (including duration). To place these results in perspective, the best prediction rates obtained by O'Connell et al. [2] averages around 35% (only 6 (scene) categories, 22 subjects, 216 photographic images). Given the disproportionate lack of content compared to photographic images, it is quite remarkable that eye-fixation sequences on sketches exhibit a predictable regularity on par with images.

In Figure 6, category-level prediction rates for 'car' and 'plant' are good since the corresponding sketches exhibit limited variations in terms of depiction viewpoint and appearance causing consistent spread in fixations compared to other categories – we refer the interested reader to view the sketch dataset of Eitz et al. [14]. Conversely, the prediction rates for certain categories ('dog', 'sheep','train') are poor since the corresponding sketches exhibit large viewpoint and appearance changes. Also, including fixation durations in saliency map computation does not help 'car' and 'plant' categories since their maps have multiple hotspots (see Figure 5) which lowers the discriminative capability induced by fixation duration.

*A. Ablative Experiments*

We also performed a series of ablative experiments over (a) the regimes – primed and unprimed[5] (b) utilizing the regimes alternately for training/testing and (c) inclusion of duration in fixation map creation. We do not include the results of these experiments in Figure 6 to retain clarity. However, we discovered the following trends:

- Models trained on primed fixations predict better and fairly above chance for test fixation sequences regardless of regime ($p < 0.005$ for primed regime and $p < 0.05$ for unprimed regime, sign-test)
- Quite a number of categories are misclassified when models are trained on unprimed regime fixation sequences. We believe this is due to confusion arising out an increased number of classes in unprimed regime (160 vs 13), the similarity among categories belonging to the same 'fixation map' group (Section 6), lack of

[5]We used fixation data of only those categories and sketches of 'unprimed setting' which were also utilized in the 'primed setting', thus making the comparisons fair.

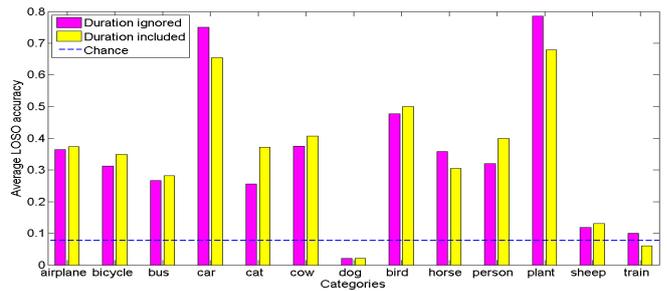

Fig. 6: Leave-One-Subject-Out predictions for categories in primed regime. All (except one) categories are predicted significantly better than chance.

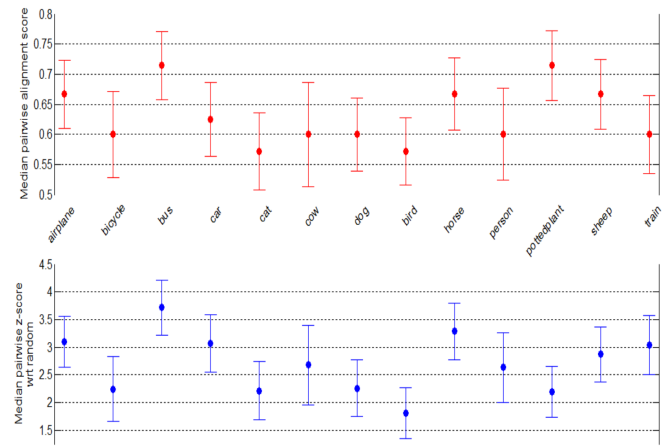

Fig. 7: Median pairwise part-sequence similarity for categories in the primed regime (red plot, top). The corresponding z-scores reflecting dissimilarity with random sequences is also shown (blue plot, bottom). Most of the categories' similarities exhibit a separation of at least two standard deviations from the corresponding mean similarity of random sequences.

  category cue as in the primed regime and the fact that the number of fixation sequences per category are smaller (38) compared to the primed regime (48).
- Inclusion of duration in fixation maps did not improve prediction performance in most of the ablative combinations.

## VIII. FIXATIONS AND SEMANTIC OBJECT-PARTS

The category prediction results in the previous section further reinforce the possibility that the fixations on sketches are not mere gestalt but are indicators of a deeper phenomenon. One hypothesis for this deeper phenomenon is the following: In the primed regime (i.e. when object category is made known), subjects fixate to regions corresponding to 'signature' semantic object-parts in a consistent manner, with the fixation visitation order reflecting the relative importance of the parts.

To verify this hypothesis, we utilized object-part contour annotations available for the sketches viewed in primed regime [18]. As a general approach, we assigned fixations to the corresponding object-part regions. This assignment of fixations to part regions entails a certain degree of ambiguity due to overlapping part annotations. To minimize this, we used a multi-stage spatial analysis algorithm to determine the assignment of fixations to part regions.

**Measuring similarity between part visitation sequences**: Assigning fixations to object part contours enables us to treat fixation sequences as part-name token sequences[6]. In general, suppose the $i$-th part token sequence for a given category $\mathbb{C}$ is given by $\mathbb{P}^{(i)} = \{P_1^{(i)}, P_2^{(i)}, \ldots P_{n_i}^{(i)}\}, i = 1, 2, \ldots n_{\mathbb{C}}$ where $n_i$ denotes the number of fixations in the $i$-th fixation sequence and $n_{\mathbb{C}}$ denotes the number of fixation sequences corresponding to category $\mathbb{C}$. To determine similarity between pairs of part-name token sequences, we used the Needleman-Wunsch algorithm [34] with a $0-1$ cost model ($Cost(P,Q) = 0$ if Part $P =$ Part $Q$, 1 otherwise) and a gap penalty of 0. The resulting similarity measure lies between 0 and 1. The median pairwise similarities for primed regime categories can be seen in Figure 7 (red-colored plot) and are fairly high (between 0.57 and 0.73).

Suppose, for a given category $\mathbb{C}$, the similarity between two fixation sequences $f_i^{\mathbb{C}}$ and $f_j^{\mathbb{C}}$ is $s_{i,j}^{\mathbb{C}}$. To verify the statistical reliability of our similarity estimation method, we first generate 100 random fixation label sequences whose length matches that of sequence $f_j^{\mathbb{C}}$. We then compute the similarity of each of the random sequences with $f_i^{\mathbb{C}}$. Suppose the mean similarity value is $\mu_{i,random}^{\mathbb{C}}$ and the corresponding standard deviation is $\sigma_{i,random}^{\mathbb{C}}$. We compute the z-score (i.e. $\frac{s_{i,j}^{\mathbb{C}} - \mu_{i,random}^{\mathbb{C}}}{\sigma_{i,random}^{\mathbb{C}}}$). We repeat this for all possible fixation sequence pairs of category $\mathbb{C}$ and compute the median pairwise random z-score for each category. The z-scores (blue-colored plot in Figure 7) indicate that the distribution of pairwise similarities between part-name token sequences is well separated from the random counterparts – most of the category-wise similarities exhibit a separation of at least two standard deviations from the mean similarity of random sequences.

## IX. A COMPUTATIONAL MODEL FOR FIXATION-BASED OBJECT PART PREDICTION

Building upon the encouraging trends in the similarity of eye-fixation sequences at fixation (Section V) and part-label level (Section VIII), we ask : Given a fixation sequence, is it possible to predict the object-part labels of each fixation in the sequence?. Such predictions could enable applications such as object-part contour annotations of freehand sketches [35] from fixations alone[7], in analogy with the object-bounding-boxes-from-eye-fixations approaches described in Papadopoulos et al. [36] and Yun et al. [33]. In addition, the additional part presence information (available for sketches annotated via fixations) can be utilized for refining results of sketch-based image retrieval approaches [37]. The consistency in fixation sequences at sketch and category level certainly suggests that the ordinal position and spatial location of a fixation within a fixation sequence has a high chance of being correlated with the object-part within whose contour the fixation falls. These observations motivate our computational model for fixation-based object part-label prediction, which we describe next.

---

[6] For example, a part token sequence from the `airplane` category could look like {fuselage,window,window,wing,-fuselage,windshield,engine}.

[7] This would be considerably less burdensome than annotating object part contours by hand.

Suppose the object category is $\mathbb{C}$. We divide the entire set of fixation sequences within $\mathbb{C}$ randomly into training and test sets. Note that in our current context, we interpret sequences to mean part-label sequences[8]. Suppose $\mathcal{P}^{\mathbb{C}} = \{\mathcal{P}_1, \mathcal{P}_2, \ldots \mathcal{P}_{n_{\mathbb{C}}}\}$ represents the part-labels for category $\mathbb{C}$.

### A. Computational Models

To model the eye fixation object-part visitation sequence, we use a Hidden Markov Model (HMM) $\lambda$. Using terminology relevant to HMMs, let us denote the observation sequence as $\boldsymbol{f}_{1:T} = (f_1, f_2, \ldots f_t, \ldots f_T)$ where $f_t \in \mathbb{R}^d$ (i.e. continuous observations) and the corresponding sequence of $\lambda$'s hidden states as $y_{1:T}$. The HMM is typically specified by three components $\lambda = (\boldsymbol{A}, \boldsymbol{B}, \pi)$.

- $\boldsymbol{A}$ is the state transition matrix such that $A_{ij} = p(y_{t+1} = j | y_t = i)$
- $\boldsymbol{B}$ is the observation model, i.e. $B_t = p(f_t | y_t)$
- $\pi$ is the probability vector of initial states, i.e. $\pi_i = p(y_1 = i)$

Let $\mathcal{F} = (f_1, f_2, \ldots f_M)$ represent the sequence of fixation features wherein $f_i$ denotes features corresponding to the $i$-th fixation. We treat $\mathcal{F}$ as observations arising from a hidden sequence $L = (l_1, l_2, \ldots l_M)$ of states. In general, the number of hidden states is a hyper-parameter. But in our case, a natural choice for hidden states is the set of part-labels $\mathcal{P}^{\mathbb{C}}$, i.e. $l_i \in \mathcal{P}^{\mathbb{C}}, 1 \leq i \leq M$.

*1) Objective:* Our objective can be stated as: *For each $l_t$, what is the most likely instantiation given the entire observation sequence $\mathcal{F}$ ?*

$$l_t^* = \underset{l_t \in \mathcal{P}^{\mathbb{C}}}{\operatorname{argmax}} \ p(l_t | \mathcal{F}) \quad (6)$$

This is referred to as pointwise maximum a posteriori (PMAP) estimate. Essentially, PMAP maximizes the posterior probability $p(l_i | \mathcal{A})$ of each individual hidden state given the *entire* observation sequence [38]. Note that the objective above is different from conventional Viterbi decoding, i.e. *What is the most likely sequence of states given the entire observation sequence $\mathcal{F}$ ?*

$$\mathcal{L} = (L_1, L_2, \ldots L_M) = \underset{(l_1, l_2, \ldots l_M)}{\operatorname{argmax}} \ p(l_1, l_2, \ldots . . . l_M | \mathcal{F}) \quad (7)$$

Our choice of PMAP is motivated by the fact that PMAP maximizes the expected number of correctly estimated states and hence provides better state estimates overall compared to Viterbi decoding [39].

*2) Fully-supervised HMM training:* Note that in our case, the observations $\mathcal{F}$ and the corresponding hidden state instantiations $y_{1:T}$ are simultaneously available for training the HMM. Thus, the training of HMM is fully supervised. In such a scenario, the maximum-likelihood training of HMMs reduces to a counting process wherein the observation, transition and initial state models can be modeled and estimated independently [40].

**Observation model:** Let $\mathcal{Q} = \mathcal{P}_i, 1 \leq i \leq n_{\mathbb{C}}$. For a given training sequence (of part-labels), we first determine the subset of labels which equal $\mathcal{Q}$. Suppose the $j$-th fixation label of a given training sequence equals $\mathcal{Q}$. We compute the corresponding ordinal factor $r_j = \frac{N_F - j + 1}{N_F}$ for the fixation where $N_F$ denotes the maximum length among all fixation sequences of the gaze data-set. The ordinal factor captures the notion that the earlier the part-label occurs in a fixation sequence, the more its relative importance within fixation sequences. This choice of ordinal factor also has the effect of normalizing for different fixation sequence lengths since the factor is always 1 for the first fixation. The feature vector $a_j$ for the fixation is constructed as $f_j = (r_j, x_j, y_j)$ where $(x_j, y_j)$ denotes the normalized coordinates of the fixation.

---

[8] By mapping each fixation to the object part within whose boundary the fixation falls, we obtain the part-label sequence.

Next, we construct a normalized non-parametric distribution $K_\mathcal{Q}$ over $f_j$s across all occurrences of the part-label $\mathcal{Q}$ in the training data. In particular, we model the distribution via Kernel Density Estimation [41] with the bandwidth automatically selected using single-dimensional likelihood-based search [42]. We repeat this procedure for each $\mathcal{Q}$ to obtain our HMM's observation model $B_t = p(f_t | y_t = \mathcal{Q})$.

**Transition and Initial state models:** In our case, the state transition probabilities $A_{ij}$ and initial state distribution $\pi_i$ where $1 \leq i, j \leq n_\mathbb{C}$ can be obtained by normalizing the appropriate part-label frequency counts from the training set sequences.

### B. Experimental Setup

For our experiments, we used $60\%$ of the category's fixation sequences selected at random to construct the per-part training models. To ensure sufficient training data, we performed data augmentation. For each reference training sequence, we generated 50 augmented sequences wherein the fixation position was randomly perturbed. During this process, we ensured that for each reference fixation, the perturbation resulted in less than 1 degree of visual angle position deviation. For each test sequence, we computed the proportion of correctly predicted part-labels (sequence prediction accuracy). The category-wise part-visitation sequence prediction accuracy was averaged over 10 trials[9]. To verify that the model's performance was better than chance, we computed predictions using randomly assigned part-labels to fixations of a sequence.

### C. Alternate models

In this section, we provide a brief overview of the alternate models we explored for the part-prediction computation model.

*1) DTW:* In the DTW model, we utilize the spatial locations of fixations. For each sequence of test fixations, we find the training fixation sequence with the optimal alignment distance $d^*$. In determining the optimality criteria, we use Euclidean metric as the base distance between fixation locations across test and train sequences. Having determined the training sequence with the smallest distance $d^*$, we determine part-label predictions for the test sequence as follows: Suppose $p$ and $q$ contain the indices of test and 'optimal-matching' train sequences respectively, as determined by the DTW procedure.

---

[9]In each trial, the training set was selected randomly.

Initially, all of test fixations $p$ are considered unmatched. Each $p_i$ is initially assigned the DTW match $q_j$. Additionally, $p_i$ is flagged as 'provisionally matched'. If $p_i$ is not matched to any other point in $q$, then the 'provisional match' becomes the final match. However, if $p_i$ participates in a subsequent match with $q_k, k > j$, then it is matched to $q_k$ provided the corresponding fixation location is closer (in Euclidean distance) to the test fixation indexed by $p_i$. Once all the matchings are obtained, the part-label prediction for each $p_i$ is obtained as the part-label corresponding to the training fixation indexed by $q_j$.

*2) RNN:* Recurrent Neural Networks (RNN)s have become a popular choice for sequence analysis in recent times, particularly for their ability to capture long-range interactions with the sequence. We evaluated different RNNs across a large variety of parameter (dropout, optimizer) and architecture (number of hidden layers) choices. The result in Figure 8 reflects the best performance out of the aforementioned choices. We also experimented with LSTMs, again with a similar variety of choices, but the results weren't too different from those obtained using RNNs.

*3) HMM-Viterbi:* As mentioned previously, conventional Viterbi decoding procedure utilizes the following objective i.e. *What is the most likely sequence of states given the entire observation sequence $\mathcal{F}$ ?*

$$\mathcal{L} = (L_1, L_2, \ldots L_M) = \operatorname*{argmax}_{(l_1, l_2, \ldots l_M)} p(l_1, l_2, \ldots . l_M | \mathcal{F}) \quad (8)$$

### D. Results for our computational model

We compared our HMM-based computation model with alternative models which are commonly used for sequence-based prediction tasks. In particular, we performed comparative evaluation with HMM Viterbi decoding (HMM-Viterbi), Dynamic Time Warping (DTW) and Recurrent Neural Networks (RNN), as described in Section IX-C.

For the evaluation procedure, the experimental setup is the same (Section IX-B). To verify that our model's performance was better than chance, we computed predictions using randomly assigned part-labels to fixations of a sequence. All the aforementioned models perform better than random. However, our PMAP-based model's performance is superior to the alternative models (see Figure 8) and distinctly better than random part-label assignment model – on average, at least $61\%$ of the sequence's labels are correctly predicted by our model across object categories.

By nature of its computations, DTW aims to stitch together optimal local alignments into a final optimal alignment. DTW-based matching typically penalizes long-range matches. Moreover, the final part-predictions are deterministically based on a 'single' best alignment. In contrast, the other methods are probabilistic in nature and can capture long-range interactions. Of these, we initially expected RNNs to outperform the HMM-based methods since they are not constrained by the Markov assumption. However, the results show otherwise. We believe one reason could be the small amount of data (even after augmentation) on a per-category basis and lack of variety therein. Another reason could be that our PMAP method maximizes the number of correctly predicted labels in expectation, which is not the case with vanilla RNNs. It would be interesting to explore PMAP-like formulations for RNNs and train them to maximize the number of correctly predicted parts. The work of Jiang et al. [22], related to learning an explicit prediction model for eye-fixation sequences, would be a worthy alternative to explore as well.

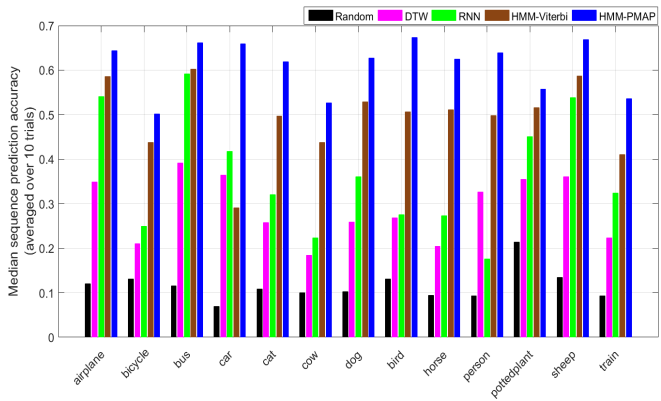

Fig. 8: Category-wise median part-visitation sequence prediction accuracies for various models. For each object category, the models are in same left-to-right order as the legend on top. Our PMAP-based model's median accuracies (blue, right-most in each group) are superior to other models and distinctly better than random part assignment based accuracies (black, left-most). Best viewed in color.

## X. Discussion

The motivation for studying eye fixations on objects arises from the works of Einhauser et al. [43] and Nuthmann et al. [44] who posit that humans more likely than not, represent and understand visual content in terms of objects (although contradictory results have been reported with non-fixation based approaches [45]). Our observations (Section VI) indicate a category-level regularity in the eye-fixation patterns on sketch objects under a free-viewing paradigm. This is in line with similar studies on photographic images [2].

fMRI-based studies have shown that the neural response of the visual system to images of a scene and line drawings depicting the scene category is virtually the same [46]. But is it also possible that eye fixation mechanisms for photos and sketches are similar? What could be the reasons they are similar or dissimilar? In our study, we have not delved into these aspects for reasons of scope. Current theories and computational models of saliency rely on features specific to photo images such as color-contrast and edge-contrast [47]. Such saliency-based explanations seem insufficient to explain our findings since such features do not exist for freehand line sketches. Therefore, at least from the viewpoint of saliency, we believe that our eye-gaze data lays the ground for uncovering connections between eye fixation patterns on objects in photographic images and their sketched versions. Such connections can help understand depiction invariant attributes and aspects of visual object representations. Combined with data from similar studies on other modalities [13], our sketch gaze data can contribute towards new insights into cross-modal (photo, brush art, line drawing) object representations [21]. Further investigation into our discovery of these interesting patterns by the community could lead to more general computational models of human visual saliency which can explain fixations regardless of depiction (photos or sketches).

Our experiments for predicting object categories (Section VI) and semantic object parts (Section IX) can be viewed as an attempt to uncover higher-level semantics from eye fixations. For example, Subramanian et al. [48] utilize eye gaze patterns to analyze semantics, albeit those related to social and affective scenes. Part-based representations of objects, semantic or otherwise, are well supported by multiple studies in neuroscience and computer vision [49]–[51]. Specifically, Taylor et al. [52] suggest that humans tap into generic concepts of objects, including linguistic propositions such as semantic object-parts, when analyzing a scene. Furthermore, studies by Palmer [53] have shown that when parts correspond to a 'good' segmentation of a figure (e.g. named-part contour annotations), the speed and accuracy of responses related to queries on figure attributes improves significantly. Combined, these observations lend support for our use of semantic object-parts as a lens to view fixation patterns.

The relationship between part-based object representations and eye fixations has been examined via imagery-based studies. In such studies, subjects are typically presented with a stimulus which is subsequently removed. The subjects are then asked to imagine or recall aspects of the stimulus previously presented to them. These studies [54] suggest that the eye fixations are an external manifestation which aim to verify an implicit, internal part-based representation for the object. Therefore, the consistency of eye fixation *order* would mirror the consistency of the representation. In this respect, the strong degree of within-category similarity among part-visitation sequences (Section VIII) for sketches is remarkable. In fact, this suggests that eye fixations on objects consistently aim to verify a part-based representation even though, unlike imagery studies, the stimulus (sketch object) has not been seen before and possibly has a different degree of detail and image quality[10]. On a related note, Xu et al. [13] perform a study which evaluates the ability of eye gaze statistics in predicting attributes of objects such as shape and size.

In addition, other studies suggest that the *order* in which the fixations happen reflects not only consistency, but the relative importance among the parts [12]. We use this observation and build upon encouraging results from part-visitation sequence similarities (Section VIII) to show that the underlying (and often visually absent) semantic-part can be predicted from the knowledge of the corresponding fixation's position within the fixation sequence and image alone ( Section IX). Our part-prediction model shows promise in this regard and we believe improved models (e.g. exploiting part-whole object semantics) can enable object-part contour annotations of freehand sketches [35] as mentioned previously (Section IX). On a deeper level, this ability to perform part-prediction extends the consistency of fixations previously seen at category (Section VI) and sketch level (Section V) to meaningful albeit implicit sub-regions (parts) of the object sketch as well.

The importance of fixation duration has been repeatedly indicated in eye-fixation and saliency literature [2], [48], [55]. In fact, Henderson [56] argues that a complete model of scene perception needs to actively take duration into account, in addition to fixation location. However, duration is typically not considered essential in most saliency prediction approaches [57]. In our case, we found that duration seemed to make a difference for the better only for predicting fixation sequence category in primed regime (Section VIII). Analysis of our fixation data reveals a negative correlation between the average fixation duration and number of fixations per sketch across categories ($-0.81$ in the primed regime and $-0.62$ in the unprimed version). Given the fixed per-sketch time budget during the study, we posit that subjects possibly trade off between number of fixations and time spent per fixation, thereby rendering duration information irrelevant for fixation-related processes such as fixation map generation.

---
[10]The internal part-based model for the object could have been constructed from a photographic image.

Our motivations for studying sketches under the 'primed' and 'unprimed' regimes were to determine whether the act of priming (a) changes the fixation sequences in a noticeable manner and (b) helps validate the consistency of implicit object part visitation via fixations. Our analysis reveals that priming helps in creation of better quality models for category prediction and fixation map generation (as attested by Figure 3).

## XI. CONCLUSION

In this paper, we have presented the first large-category scale exploration of eye-fixation data on freehand sketches. As a result of our study, an eye fixation database for 3904 hand-drawn sketches across 160 visual object categories has been obtained for the benefit of the community. Analysis of data from our user study has shown that eye-fixations on freehand sketches are not mere gestalt – they are strongly conditioned by the categorical aspect of sketch content (object). In fact, this conditioning is strong enough for the object category to be predicted from the fixation sequence alone. Even more dramatically, our results show that the sequencing of fixations corresponds to an implicit sequence of parts that constitute the object although the parts themselves may not be delineated in the sketch stroke data. In our work, we have also shown how the consistency in visitations can be used to build object-specific computational models capable of predicting the semantic object parts which underlie fixations.

More broadly, our sketch object eye-fixations data lays the ground for uncovering connections between eye fixation patterns on objects in photographic images [36] and their sketched versions. Such connections can help understand depiction invariant aspects of visual object representations and with data from similar studies on other modalities, provide new insights into cross-modal (photo, brush art, line drawing) object representations [58].

To the best of our knowledge, current theories and computational models of saliency, which implicitly assume photographs or real-world scenes as input, seem insufficient to explain our findings. Therefore, further investigation into our discovery of these interesting patterns by the community could lead to more general computational models of human visual saliency which can explain fixations regardless of depiction (photos or sketches).

## ACKNOWLEDGMENT

The authors would like to thank Prof. Veni Madhavan (Indian Institute of Science), for access to SMI Eye Tracker equipment. The authors would also like to thank NVIDIA for their donation of Tesla K40 GPU.